\documentclass[runningheads]{llncs}
\usepackage{graphicx}
\usepackage{epstopdf}
\usepackage{amsmath}

\begin{document}
\title{Representing Sets as Summed Semantic Vectors}
\titlerunning{Sets as Semantic Vectors}

\author{Douglas Summers-Stay\inst{1} \and
Peter Sutor\inst{2} \and
Dandan Li\inst{1}}
\authorrunning{D. Summers-Stay et al.}

\institute{U.S. Army Research Laboratory,  Adelphi, MD 20783, USA \and
University of Maryland, College Park, MD 20742
\email{douglas.a.summers-stay.civ@mail.mil}}

\maketitle             
\begin{abstract}
Representing meaning in the form of high dimensional vectors is a common and powerful tool in biologically inspired architectures. While the meaning of a set of concepts can be summarized by taking a (possibly weighted) sum of their associated vectors, this has generally been treated as a one-way operation. In this paper we show how a technique built to aid sparse vector decomposition allows in many cases the exact recovery of the inputs and weights to such a sum, allowing a single vector to represent an entire set of vectors from a dictionary. We characterize the number of vectors that can be recovered under various conditions, and explore several ways such a tool can be used for vector-based reasoning.

\end{abstract}

\section{Introduction}

Representing concepts as high-dimensional vectors for reasoning is one of the most powerful ideas to come out of biologically-inspired computing. Unlike purely symbolic approaches, high-dimensional vectors have a structure that is rich enough to model the subtle and complex associations that concepts in memory have with one another, allowing generalization, analogy, and the combination of ideas.

There are various approaches to generate such high-dimensional vectors, including modeling semantic primitives with random vectors and using the weights of artificial neural networks. One of the most successful (in terms of immediate applications) sources of vectors has come out of the statistical linguistics community: creating vectors for words by ensuring that words which share a similar context in text corpora are mapped to similar vectors. Throughout this paper we will work with such distributional semantic vectors, but the ideas here can be applied to any architecture that uses high-dimensional vectors to represent meaning. As such we will usually refer to these vectors as representing words or concepts, but without intending to imply that distributional semantic vectors are themselves complete models of the meaning of words or concepts. 

An operation common to many such approaches is vector addition or averaging to represent the meaning of a set of vectors. The meaning of this vector is in some way between the meanings associated with the vectors that make it up. For example, the word vector ``astronomer" averaged with the word vector ``physicist" will typically result in a vector close (in a suitable metric) to the word vector ``astrophysicist." 

However, this has generally been treated as a kind of summarization of meaning that inevitably loses track of the vectors that make up the sum. It would be convenient to be able to reverse this process: to take a vector which is a weighted sum of meaningful vectors, and find what the weights and inputs were. To a certain extent this is possible simply by looking for the nearest neighbors of the averaged vector in the dictionary. Because of inherent properties of high-dimensional Euclidean spaces and the small size of the dictionary compared to the size of the space, an averaged vector which is the mean of two word vectors will almost always be closer to these two word vectors than to any other word vector. But for more than a few words averaged together, the ability to recover words by looking at near neighbors fails rapidly. In this paper, we show how sparse vector decomposition techniques can be applied to overcome this limitation of summed vectors. This allows them to be directly useful for many kinds of tasks which would have been impractical in a vector-based representation previously. To our knowledge, this is the first use of such sparse decomposition techniques for recovering exact weights for individual word vectors from a vector sum.

What we show in this paper is the following surprising result: a single summed vector can be used to represent the entire set of words from which it was derived, as long as the set is smaller than the limits outlined in section \ref{experiments}. Rather than thinking of a summed vector as representing a single meaning, we can think of it as picking out an entire set of meanings. For example, the Wikipedia page ``list of fruits'' lists 90 different fruits. A single 600-dimensional vector\footnote{A vector that represents a category as an average is often very close to the vector representing the word for the category. In the case of the fruit category vector discussed here, the word 'fruit' is the closest word in the dictionary to the averaged vector. \cite{Kommers} suggests that this is due to the fact that the hypernym 'fruit' is associated with each of its hyponyms, but only some of the hyponyms are associated with each other.} can be used to recall this entire list, provided each fruit is assigned a vector in the dictionary. Beyond that, each fruit on the list can be assigned a weight that it is possible to recover exactly.

\section{Related Work}

Pentti Kanerva's ``hyperdimensional computing'' is mainly concerned with large binary vectors, but also can include floating-point valued vectors.\cite{Kanerva} Kanerva writes, \begin{quote}
``A set of vectors can be combined by componentwise addition, resulting in a vector of the same dimensionality... [T]he arithmetic-sum-vector is normalized, yielding a mean vector. It is this mean-vector that is usually meant when we speak of the sum of a set of vectors. The sum (and the mean) of random vectors has the following important property: it is similar to each of the vectors being added together. The similarity is very pronounced when only a few vectors are added and it plays a major role in artificial neural-net models. \textit{The sum-vector is a possible representation for the set that makes up the sum} [emphasis added].'' 
\end{quote}
To recover elements of the set, the paper recommends searching the dictionary for the nearest neighbor, subtracting it off, and searching for the nearest neighbor of the remainder. He notes that only small sets can be decomposed in this way.\footnote{Kanerva typically works with 10000-dimensional vectors, which would allow this method to work for somewhat larger sets than the 25-600 dimensional vectors discussed in this paper.} These ideas are explored further in the subfield called Vector Symbolic Architectures.\cite{Levy}

In the Hierarchical Temporal Memory model introduced by Jeff Hawkins, concepts are represented by sparse high-dimensional binary vectors, and the ``union" of concepts is formed by applying the logical OR operation to these. If the logical AND operation is applied to a vector and this union and the original vector is returned, it belongs to the set with a probability that is easily calculated.\cite{Taylor} This method gives no way of weighting concepts in the sum, however, and is not easily applied to floating-point valued vectors.

The notion of representing the meaning of a set by the simplex whose vertices are the set members in a semantic vector space was explored by Peter Gardenfors\cite{Gardenfors}. A practical application of this with vectors derived from the weights of a neural network was discussed in \cite{Bechberger}.

Dominic Widdows and Trevor Cohen discuss a logic of projection operators in vector spaces\cite{Widdows}. In their system, "Each projection operator projects onto a (linear) subspace; the conjunction of two operators projects onto the intersection of these subspaces; their disjunction projects onto the linear sum of these subspaces; and the negation is the projection onto the orthogonal complement." When these subspaces have the restriction that weights on semantic vectors must sum to one and be positive, they are cut down to simplices and the intersection and union operations he introduces become those discussed in this paper.

Our previous work discusses the possibility of representing a concept by a summed vector of synonyms of the concept\cite{Summers-Stay2} and representing the relation between concepts as a vector which can be decomposed into a chain of reasoning\cite{Summers-Stay} as discussed in section \ref{Deductive Logic}. In that work we used a fixed $\lambda$ parameter for LASSO and didn't attempt an exact recovery of the associated weights.

\section{Decomposition of Vector Sums}

Distributional semantic vector spaces represent words as (typically normalized) high-dimensional Euclidean vectors. A set of word vectors is often represented by a sum or average of each word in the set. We will refer to this as a \textit{summed vector} or an \textit{averaged vector} (if it has been normalized) as opposed to the \textit{word vectors} listed in the \textit{dictionary} (a mapping between the strings for words and their associated vectors). The problem of finding a weighted sum of vectors that adds to another vector is a linear regression problem, equivalent to solving a system of linear equations, where the weights are the solutions to each variable and the word vectors are the columns of coefficients on each respective variable. When the number of word vectors in the dictionary is larger than the dimensionality of the vectors, the solution is not unique. In order to choose from among these solutions, we impose the restriction that the weights on most words will be zero (sparsity). 

This is known as a sparse sum problem, and can be solved with tools such as the Least Absolute Shrinkage and Selection Operator (LASSO)\cite{Tibshirani}, which selects a few vectors to have non-zero weight in order to approximate the sum.

LASSO balances sparsity against exactness in finding sparse sums with a parameter, $\lambda$. It can be difficult to choose the correct lambda, so we use a screening method called DPP (Dual Polytope Projection)\cite{Wang} to efficiently test over the full range of $\lambda$ values from 0 to 1, and gather all the candidate non-zero weighted vectors from this range. With a dictionary size less than or equal to the dimensionality of the vectors, it is possible to simply solve the resulting system of linear equations exactly. That is, if it is possible to narrow down the list of possible non-zero weighted vectors to no more than the dimensions of the embedding space, all the weights and associated word vectors can be exactly recovered. Because of this, we choose the $n$ vectors given the highest weight, where $n$ is the dimensionality of the vectors, and solve exactly with these vectors as the dictionary. As long as the correct vectors are included among these $n$ vectors, the exact weights will be recovered.

There is good reason to believe that the brain makes use of some kind of sparse decomposition in order to make sense of complex inputs\cite{Beyeler}. It has been speculated that this could be achieved through competition among neural units in a winner-take-all architecture\cite{Coultrip}. Others have found that sparsity can be achieved by appropriate thresholding\cite{Rozell}. Note that in the experiments in this paper, the individual word vectors are \textit{not} sparse. The sparsity is in which words are chosen from the dictionary.

\begin{table} \label{table1}
\caption{Screening methods outperform nearest-neighbor and LASSO for recovery of word vectors from an averaged vector. In this experiment, a dictionary of 100,000 300-dimensional vectors was used.}\label{tab1}
\begin{tabular}{|l|r|}
\hline
method & number of word vectors recoverable\\
\hline
\hline
nearest-neighbors &  2\\
\hline
LASSO with $\lambda=.02$ & 5 \\
\hline
DPP-screening with $\lambda$ ranging from 0 to 1 &36\\
\hline
\end{tabular}
\end{table}

\section{Experiments: Recovery of Weights and Word Vectors}\label{experiments}

When trying to recover sparse solutions, there is a phase transition between problems where the solution can be recovered exactly, and problems where this rarely is possible. Problems whose sparsity (non-zero weights / dimensionality) is low enough and whose undersampling ratio (dimensionality / dictionary size) are high enough can be shown to be exactly solvable.

\begin{figure}
\centering
\begin{minipage}[t]{.45\textwidth}
  \centering
  \includegraphics[width=1\textwidth]{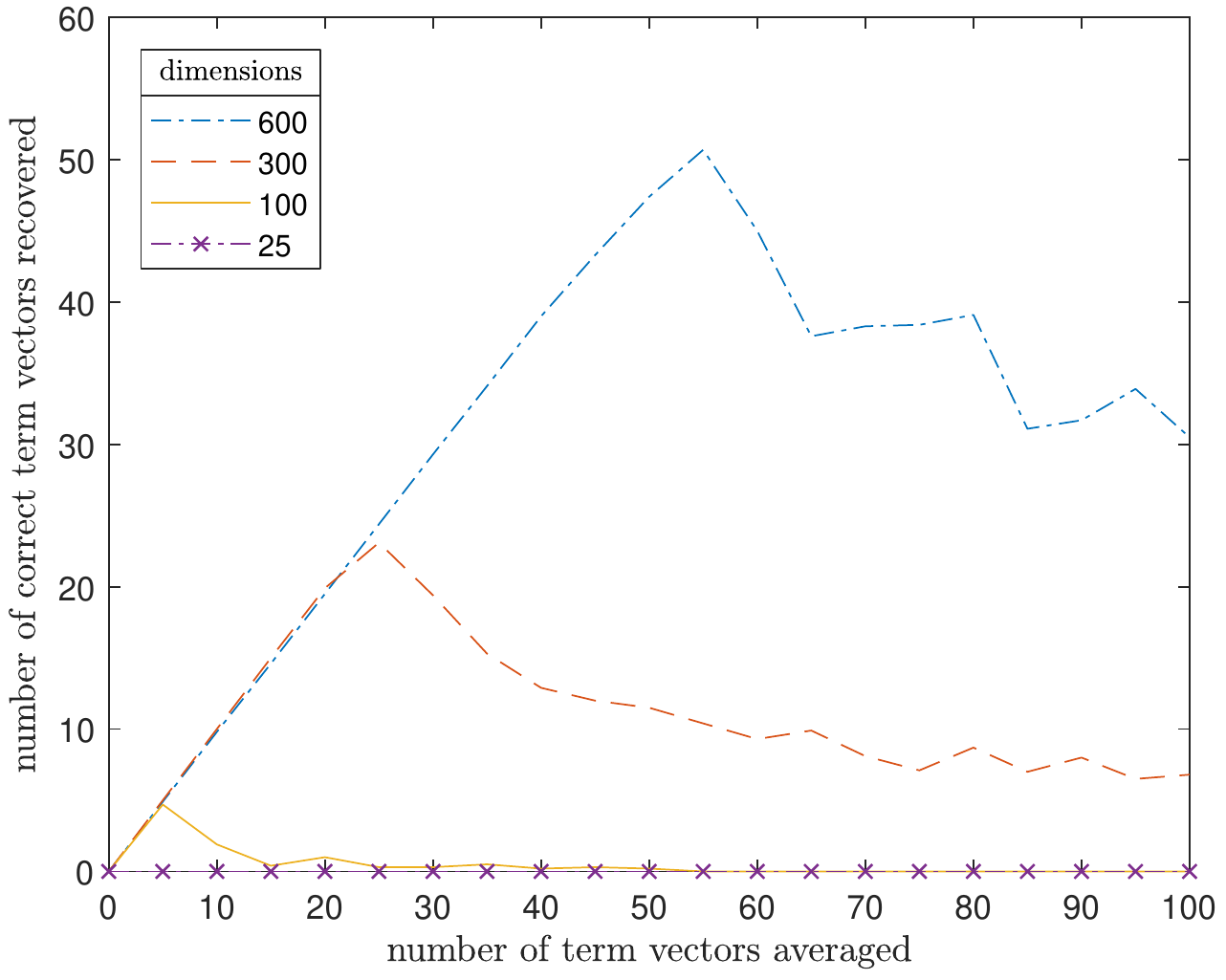}
\caption{Decomposition of averaged vectors with a dictionary of size 538116 with varying vector dimensions.} 
  \label{fig1}
\end{minipage}\qquad
\begin{minipage}[t]{.45\textwidth}
  \centering
\includegraphics[width=1\textwidth]{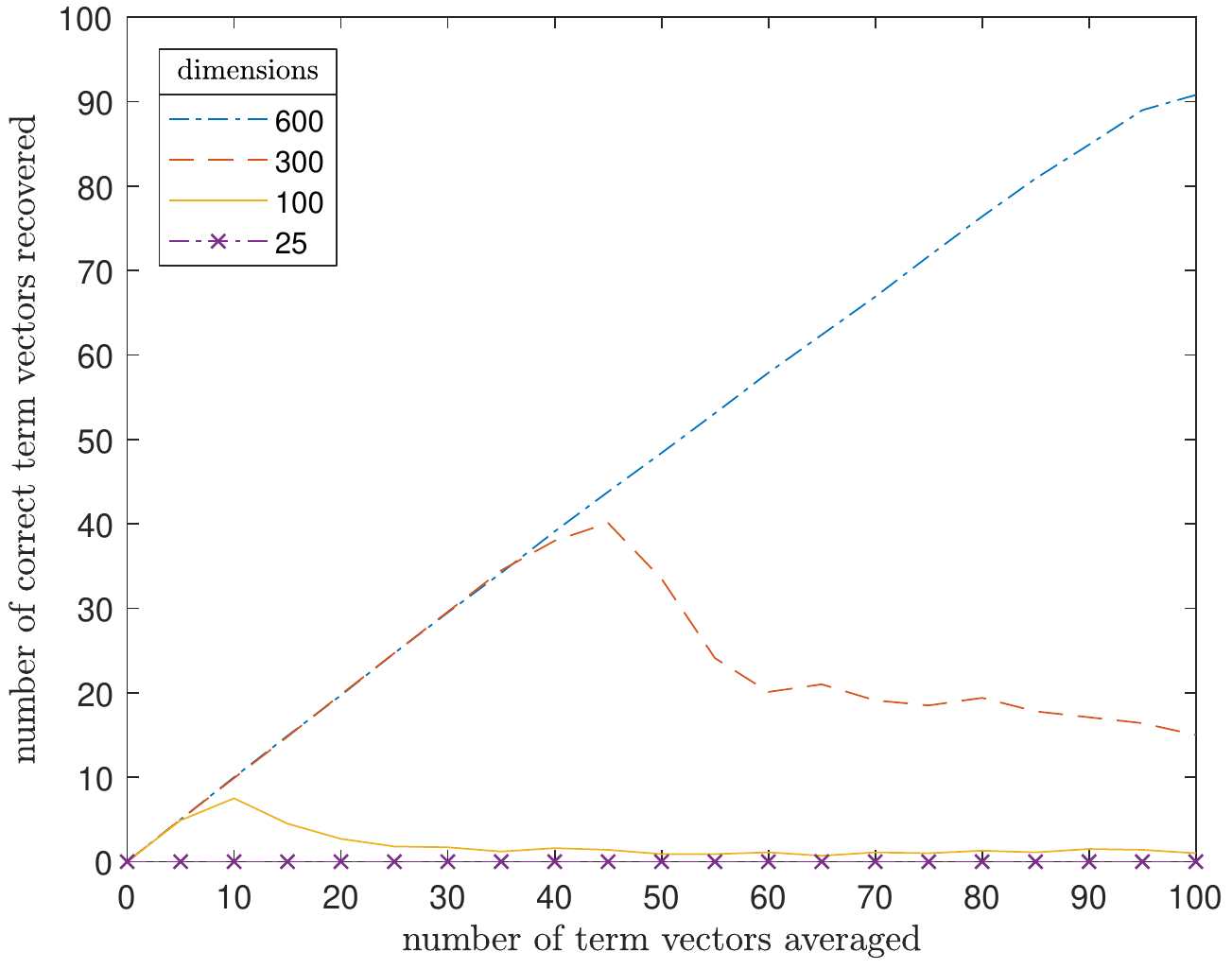}
\caption{Decomposition of averaged vectors with a dictionary of size 10000 with varying vector dimensions.} \label{fig2}
\end{minipage}
\end{figure}

The theoretical probability of a solution being recoverable has to do with the dual of the LASSO problem. It can be calculated by the following formula (see \cite{Donoho} for a derivation): $$\frac{f_k(AC^N)}{f_K(C^N)}$$ where $f_k$ is the number of k-dimensional faces on a polytope, $C^N$ is the cross-polytope\footnote{The cross-polytope is a generalization of the square and octahedron to higher dimensions.} with $N=$ dictionary size, and $A$ is the dictionary matrix which is $n \times N$. For example, for a dictionary size of 10000 and a dimensionality of 100, 300, and 600, the number of terms that should be recoverable in \textit{most} cases is 11, 41 and 97 respectively. These results match the experimental results in figure \ref{fig2} fairly well. According to the theory, below 3, 11, and 25, respectively, \textit{all} weights can be recovered perfectly for vectors of these dimensions. The word vectors in the n-dimensional models used in this paper are all well modeled by an n-dimensional normal distribution with a mean of zero and are approximately in general position, which means these theoretical bounds should be applicable.\footnote{A calculator for the 100\% solvable and 50\% solvable cutoff in terms of sparsity and undersampling ratio can be found at \url{people.maths.ox.ac.uk/tanner/polytopes.shtml}.} In these experiments we varied four parameters: 

\begin{enumerate}
\item Dimensionality of the word vectors
\item Number of vectors averaged
\item Semantic similarity of the vectors being averaged
\item Amount of noise
\end{enumerate}

\begin{figure}
\centering
\begin{minipage}[t]{.45\textwidth}
  \centering
\includegraphics[width=1\textwidth]{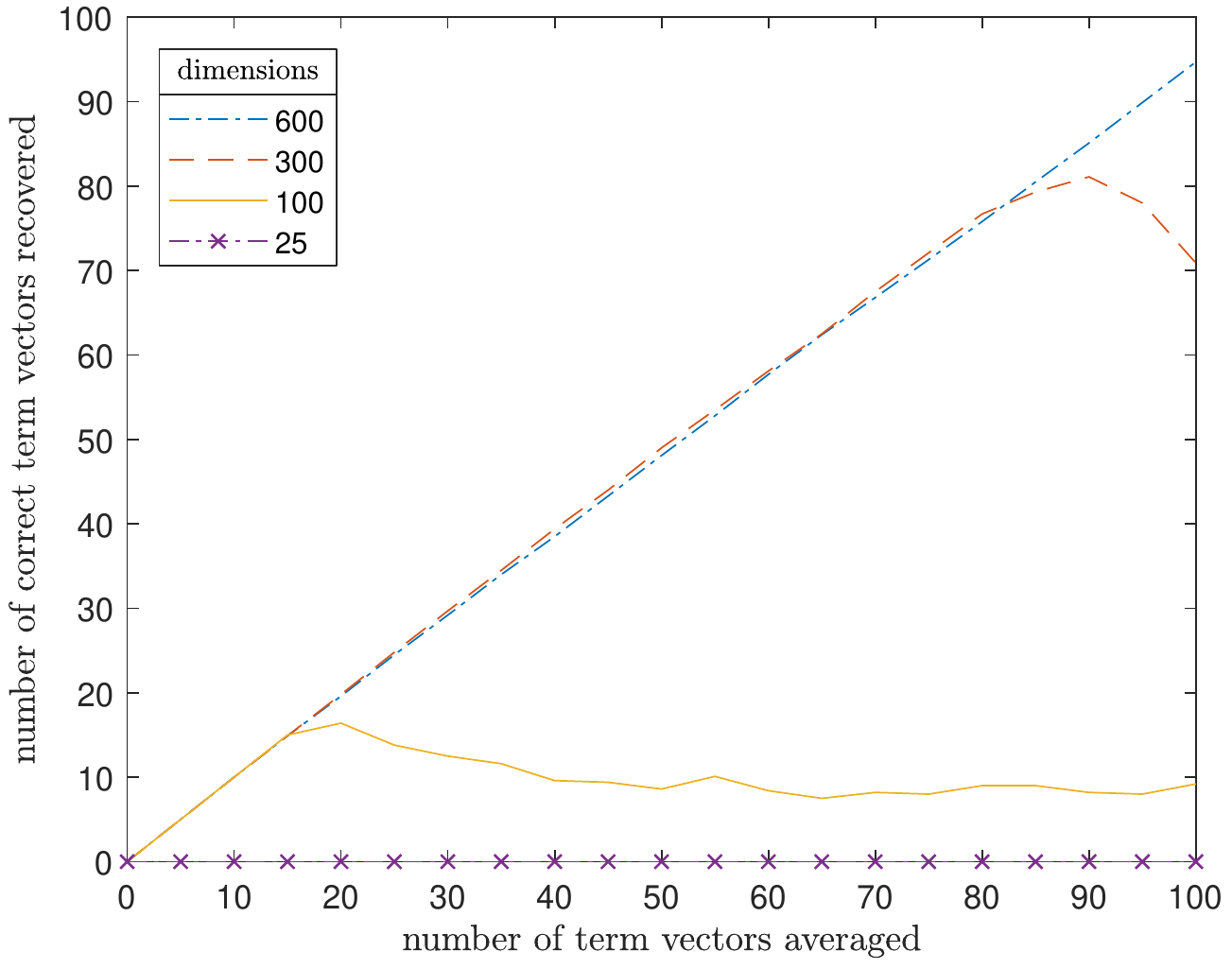}
\caption{Decomposition of averaged vectors with a dictionary of size 1000 with varying vector dimensions.} \label{fig3}
\end{minipage}\qquad
\begin{minipage}[t]{.45\textwidth}
  \centering
\includegraphics[width=1\textwidth]{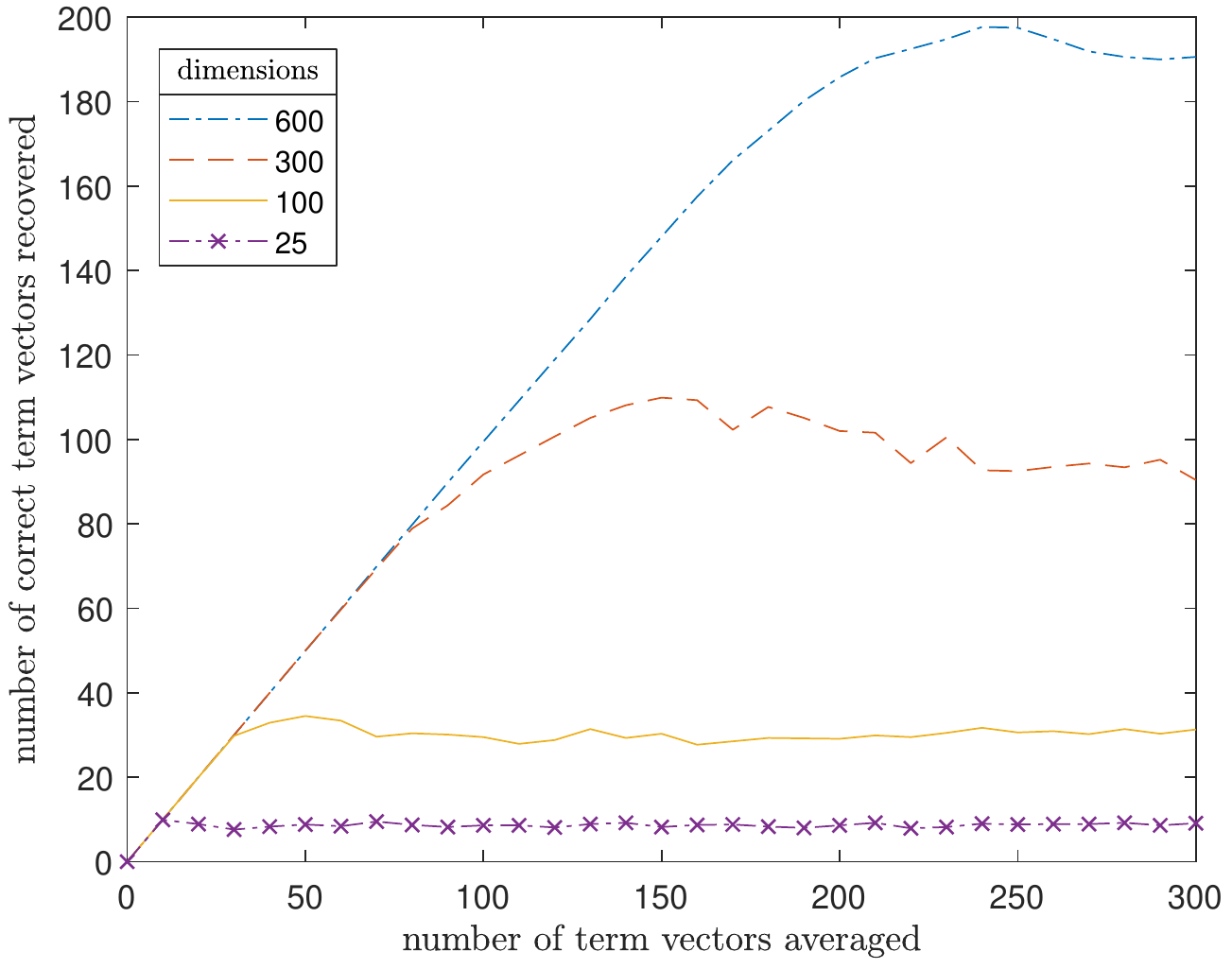}
\caption{Decomposition of averaged vectors (which are all near a common vector) with a dictionary of size 538116, with varying vector dimensions. Compare to figure~\ref{fig1}.} \label{fig4}
\end{minipage}
\end{figure}

We used publicly available word embeddings from the Bar-Ilan University NLP lab\cite{Melamud}.\footnote{We also verified that results reported for 300-dimensional vectors were similar for Mikolov's 300-dimensional word2vec embeddings trained on the Google News corpus.} Word embeddings created with a window size of 5, of dimensions 25, 100, 300, and 600 were tested. This data was normalized before the experiments. The experiments were run ten times and the results averaged to emphasize the trend rather than the variability.
In general, figures~\ref{fig1}-~\ref{fig3} show that a smaller dictionary size and a larger dimensionality of the embedding space lead to a larger number of vectors which can be recovered. This can be thought of in terms of density of terms in the embedding space: when the density is low, the number of terms that can be recovered is high.

\begin{figure}
\centering
\begin{minipage}[t]{.6\textwidth}
\includegraphics[width=1\textwidth]{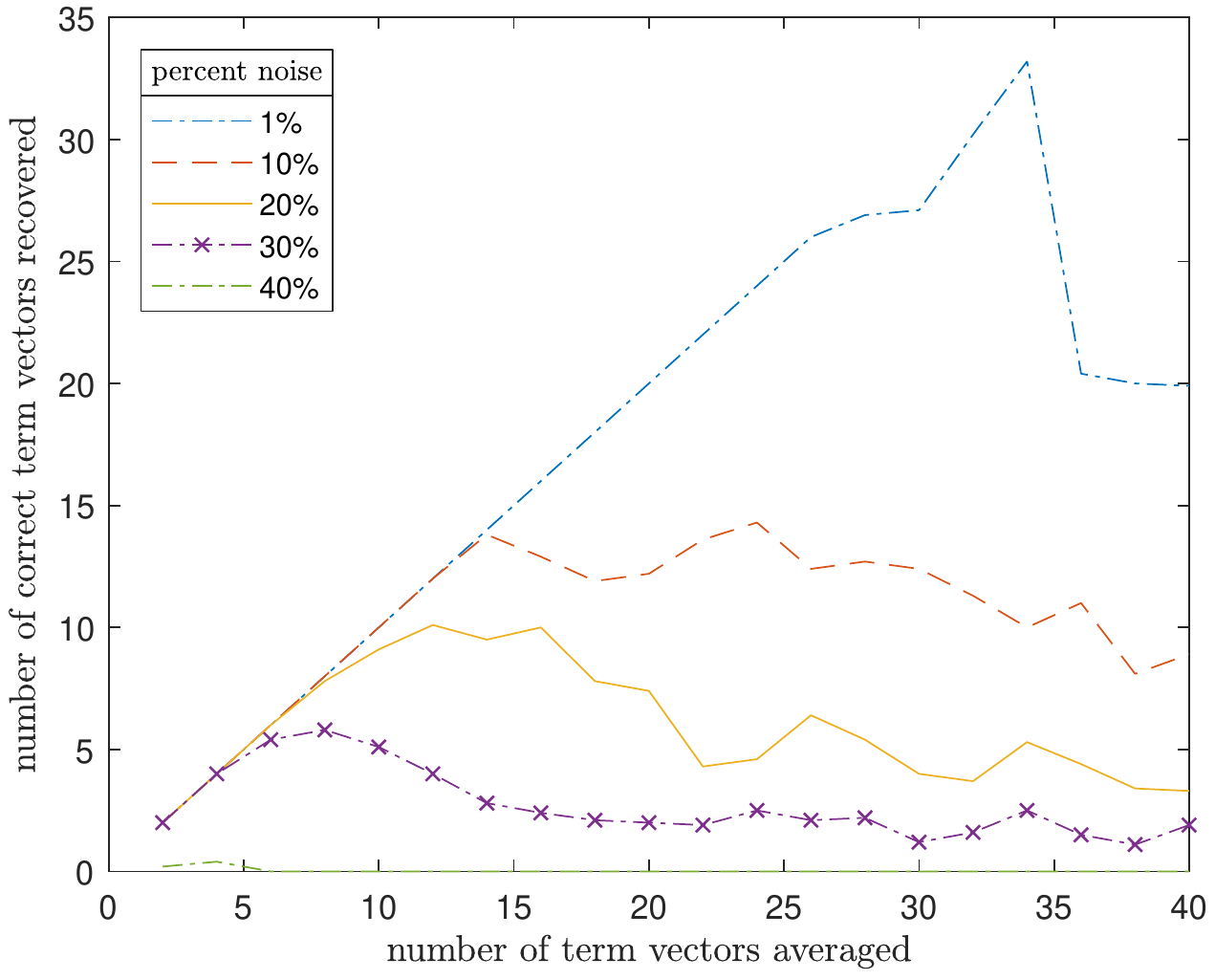}
\caption{Decomposition of averaged 300-dimensional vectors with a dictionary of size 538116, with varying amounts of noise.} \label{fig5}
\end{minipage}
\end{figure}

However, in most real uses, the vectors grouped in a set will have something in common. To simulate this situation, we chose one word at random and chose half of its nearest neighbors at random. These words are all clustered together in the embedding space, but are mixed together equally with words that are not part of the average. In this situation we were able to recover a much larger number of words: for the same dictionary size and vector dimensions which was able to recover around 20 words in the general case (see Figure~\ref{fig1}), we could recover more than 90 weights and word vectors (see Figure~\ref{fig4}). 

The decomposition process is also robust to the presence of noise, as seen in Figure~\ref{fig5}. This is especially important in architectures that mimic the noisy environment of the brain. Vectors which do not appear in the dictionary would also have a similar effect to noise, and some of the operations that one might wish to apply to a vector, such as looking for analogous concepts, can be treated as finding the true answer plus an unknown noise vector.

\section{Applications of Summed and Averaged Vectors}

The decomposition process\footnote{In order to perform a decomposition using DPP, a summed vector must be first be normalized. It can be de-normalized after the decompostion.} allows us to map vectors in a consistent way from the original vector space (with 300 or so dimensions, whose basis vectors are uninterpretable) to a new sparse overcomplete basis made up of the words in the dictionary. The canonical example from \cite{Mikolov}: $$X = 1.0 \times king - 1.0 \times man + 1.0 \times woman$$ can be thought of as a vector in this new basis. When it is mapped back into the original basis, it can be found to be nearby the vector for $queen$. As this is simply a change in basis, vector addition and scalar multiplication on vectors in one basis still hold in the other basis. 

There are several different interpretations we can give such a vector.
\begin{enumerate}
\item \textbf{A weighted semantic average}. The most direct interpretation is that an averaged vector represents a word semantically between the words which it is the weighted average of. If the word ``nashi" is not in the dictionary, we might represent it by a vector halfway between apple and pear (two fruits it resembles and is related to.) Generally speaking, all the properties of something ``semantically between" two other things can be expected to share the properties of one or the other, or have a property which is itself a blend between the properties of the two. A nashi, for example, has a yellow-green color and a flavor similar to a pear but a crunchiness and shape closer to an apple. This isn't always true -- some properties of an object semantically between $a$ and $b$ may be very different from either $a$ or $b$ -- but the exceptions have to be learned as exceptions, while this can be considered the default. This interpretation requires that all weights are positive and sum to one.

\item \textbf{A multi-set}. The summed vector can represents all of the word vectors that make it up, with integer weights above indicating that there is more than one copy of the concept in the multi-set. To add something to the multi-set, we add its semantic vector to the summed vector, and to remove a copy of it from the multi-set we subtract it. This interpretation requires that all weights are positive integers, although generalizations of multi-sets with negative multiplicities could also be used. 

\item \textbf{A set}. To represent a set rather than a multi-set, weights above one must be thresholded to one after anything is added to the set. Intersection of sets A and B can be accomplished by decomposing the summed vectors $\vec{a}$ and $\vec{b}$, finding the intersection of the sets of word vectors, and creating a new summed vector from the set of words $\{A \cap B\}$. To form the set union, one can decompose $\vec{a}$ and $\vec{b}$, and form a new vector as the sum of all the word vectors in either $\vec{a}$ or $\vec{b}$. 

However, the decomposition operation is more computationally expensive than addition of vectors, so it is desirable to minimize its use by performing the decomposition after a series of intersection or union operators have been applied. Suppose we have three vectors $\vec{a}$, $\vec{b}$, and $\vec{c}$, representing the sets $A$, $B$, and $C$. We assign each word vector in $\vec{a}$ equal weight $i$, each word vector in $\vec{b}$ weight $j$, and in $\vec{c}$ weight $k$. We form a new vector $\vec{d}=\vec{a}+\vec{b}+\vec{c}$. Then we decompose $\vec{d}$. The weights of the resulting vectors tell what part of the Venn diagram the vector falls into. Those word vectors with weight $i+j+k$ fall into the intersection of all three sets, those with weight $i+j$ fall into the region $A\cap B \cap \neg C$ (and similarly for the other subsets), and those with any positive weight fall into the union $A \cup B \cup C$. These weights will be incorrect if the bounds in the experiment section are exceeded, but the general principle that the most highly weighted word vectors in the sum belong to the intersection of the most sets being combined will still be approximately true.  

The symbol $\top$ can be used to represent the sum of all the vectors in the dictionary, $v_1+...+v_n$. The set $\neg A$ can then be written $\top - (\vec{a_1} + ... + \vec{a_n})$. The union of any positive term $\vec{v}$ with $\top$ is $\vec{v}\cup\top=\top$, while the intersection is $\vec{v}\cap\top=\vec{v}$. The set $\{B\cap\neg A\}$ can then be calculated as $B-B \cap A$.\footnote{Widdows \cite{Widdows} suggests that $\{B \cap \neg A\}$ be calculated by projecting $\vec{b}$ onto the space perpendicular to $\vec{a}$. This is equivalent to adding $-w*\vec{a}$ to $\vec{b}$, for some particular positive $w$, but this $w$ will be higher than is strictly necessary to reduce the components in $a\cap \neg b$ to zero, if the vector $b \cap \neg{a}$ is not already orthogonal to $a \cap \neg b$. This may be desirable, depending on how the result is to be interpreted. In practice, though, the nearest neighbors of a vector from the dictionary are surprisingly robust to changes in the weight of that vector, so finding the correct value for w may not be critical for many applications.} If needed, $\top$ could be included as a word in the dictionary.

\item \textbf{A fuzzy set}. When we know that our knowledge in an area is complete, performing intersections by taking the literal set intersection given above is correct, and won't introduce any error. But in most real-world applications, we are less than completely certain. Should strawberry preserves be included in the intersection of the set of fruits with the set of things that are sweet? What if we are talking about nutritional content rather than botany? It would be impossible to quantify all of these issues directly except for very narrow topics. For many applications, though, it is helpful to have a graded notion of set intersection and union. The purpose of embedding in a vector space is to take advantage of this ``approximate knowledge of many things" that is implicitly available in the structure of the embedding. 

Fuzzy set theory allows inclusion in a set to be a matter of degrees, which is also known as graded classification\cite{Kalish}. If weights between zero and one are allowed, the set interpretation becomes a fuzzy set interpretation. Prototypical examples of a class would be given the highest weight, while more marginal members would be given lower weight. Such weights could be assigned by taking a survey of usage (perhaps $15/20$ respondents felt that $strawberry\_preserves$ should belong to this set, for example, so it would be assigned a weight of $.75$ in the pre-normalized sum) or, if such information is unavailable, including terms near to the known members with weight that decreases as distance from the known core increases according to some monotonically decreasing function.  

In fuzzy set theory, the notions of intersection and union are generalized to handle fractional set inclusion. To calculate the intersection of fuzzy sets $F$ and $G$, we use the function $$F \cap G= min(w_{F1},w_{G1})\vec{v_1} + ... +min(w_{Fn},w_{Gn})\vec{v_n}.$$ The union is $$F \cup G= max(w_{F1},w_{G1})\vec{v_1} + ... +max(w_{Fn},w_{Gn})\vec{v_n}.$$ These functions can also be used for intersection and union of standard sets when the weights are ones and zeros.  

\item \textbf{A probability distribution}. Weights here represent a probability distribution over the averaged word vectors. The weights are interpreted as probabilities, and the distribution of weights must be positive and sum to one. Taken this way, the vector $(.6 * \vec{apple} + .4 * \vec{pear})$ would mean a probability of .6 that the object is an apple, and .4 that it is a pear. If we have some function that maps the vector $\vec{apple}$ to $\vec{round}$ and $\vec{pear}$ to $\vec{pear\_shaped}$, and the function is locally linear, then it would map this vector to $(.6 * \vec{round} + .4 * \vec{pear\_shaped})$, showing how probabilities about entities can be carried over into probabilities about their features.

If the probability distributions are independent, we can calculate a vector representing the probability of $(A\ or\ B)$ by taking the sum of averaged vectors $\vec{a} + \vec{b}$ and re-normalizing (similar to set union above). The probability of $(A\ and\ B)$ can be calculated by decomposing $\vec{a}$ and $\vec{b}$, taking the element-wise product (that is, multiplying the probabilities of each word in $A$ with the probability of the same word in $B$), re-normalizing, and creating a new averaged vector from the new renormalized weights (which amounts to set intersection in the case of equal weights). 

\item \textbf{An information structure}. Because the weights are recovered by decomposition as well as the vectors, some information about additional structure on the set can be carried by the weights themselves. As a simple example, one can assign a weight of 1 to the vector for the first word in a sentence, 2 to the second word, and so on. When the weights are recovered, the order of the words is also recovered.\footnote{A slightly more sophisticated weight coding scheme would be needed to handle sentences which may contain repeated words, such as using a random floating-point number to represent each position and using a lookup table for potential sums of these.} Another possibility is weighting the object of the sentence more heavily than the subject (or vice-versa) to distinguish them. Perhaps even the entire tree structure of a parsed sentence could be encoded in the weights. These ideas are not yet fully developed, and desirable properties of semantic nearness between sentences will not be preserved by such arbitrary weighting schemes.
\end{enumerate}

One challenge is that these various interpretations interfere with one another: if we have some averaged vectors representing semantic averages and others representing probability, it is impossible to tell them apart, and operations on them will tend to mix these interpretations together. 

\subsection{A Class Simplex}

An averaged vector is decomposed into its component word vectors by solving a system of linear equations, whose rows are the equations and whose columns are made up of the word vectors. This is equivalent to finding the barycentric coordinates of the averaged vector within a simplex whose vertices include the word vectors. In the case of the derived fruit vector discussed above, for example, the vertices of this simplex would be the vectors for the original list of fruits. Instead of 90 individual points, we have a 90-dimensional region that represents the set of fruits. Using the first interpretation (semantic betweenness), each point within this space represents some imaginable fruit.  Although we may not be able to name all of these possible fruits, we would still likely recognize each as belonging to the class ``fruit" and any properties of this fruit we would expect to be shared by some of the other fruits. Depending on the weights, the fruit represented in this way can be more or less similar to any subset of fruits. Gardenfors argues\cite{Gardenfors} that all natural concepts are convex regions of conceptual space. Although his ideas for how to create the dimensions of the embedding space are different, the principle is the same: if the concepts at all vertices of a face of a simplex share a particular property, then we can assume that any point within that face also shares that property.

We should not, however, expect to find any word vectors inside this region among the words in the dictionary (except those forming the vertices) because it takes up such a tiny fraction of the overall space. If our knowledge is incomplete, there may be word vectors that should belong to the class but haven't been added yet. We can measure the distance from any point to the nearest surface of the simplex, and points nearby can be considered as likely candidates for belonging to the class. The projection of the point onto this nearest point of the simplex, in barycentric coordinates, tells which words from among the set it is most semantically similar to, and with what weights.

Measuring distance to the nearest point on the surface of the simplex is more accurate for classification than distance to the class mean. We created 3734 classes of size between 30 and 300 by taking triples from conceptnet\cite{conceptnet} with the same relation and second term, but differing in their first term. For each class, the mean $L_2$ distance from a randomly selected 300-dimensional vector from this list to the mean of the rest of the vectors in the class was .94 with a standard deviation of .05, while the mean distance from the vector to the simplex formed by the rest of the vectors in the class was .88 with a standard deviation of .08. This indicates that distance to the simplex can discriminate somewhat between vectors belonging to a class and vectors the same distance from the mean which do not belong to the class. This is due to the shape of the class distribution, which is flattened in some dimensions. A similar effect was used in \cite{McGregor} where only the most significant dimensions were used to measure distance to the center of a class.

\subsection{Deductive Logic} \label{Deductive Logic}
To find a chain of reasoning in propositional logic, one combines statements about propositions such as $A \rightarrow B$ and $B \rightarrow C$ to conclude $A \rightarrow C$. These propositions can be thought of as picking out a set of terms for which the proposition holds, so a summed vector can be used to represent a proposition. The statement $A \rightarrow B$ can be considered as an instruction to replace A with B, which can also be represented as the summed vector $-\vec{a} + \vec{b}$. (This is closely related to the "logical space" system described in \cite{Westphal}.) We can create a knowledge base of such statements stored as vectors, and use that knowledge base as a new dictionary over which we can perform decomposition. To test whether $ A \rightarrow D $, we form the vector $-\vec{a}+\vec{d}$ and see whether it can be decomposed into vectors from this knowledge base, such as

\begin{equation}
(-\vec{a}+\vec{b})+(-\vec{b}+\vec{c})+(-\vec{c}+\vec{d}) = (-\vec{a}+\vec{d})
\end{equation}

If the sets A,B,C, and D above are all represented by the summed vector of the elements belonging to the set, the quantities $(-\vec{a}+\vec{b})$, $(-\vec{b}+\vec{c})$, and so forth will all be a positive sum of word vectors, since implication must alway take a set to a superset (if Socrates is in the set of all men, and we are given the fact that $\{men\} \rightarrow \{mortals\}$, Socrates must belong to the set of mortals). This is a little disingenuous, since if we already knew precisely which elements belong to each set there would be no need to perform deductive reasoning at all. However, even if the intermediate sets are represented by some other vectors, the chain can still be found because these all intermediate steps cancel out in the end.

When the knowledge base is embedded in a semantic vector space it can go beyond what can be deduced from the premises in purely deductive reasoning into what Charles Pierce called ``ampliative reasoning." \cite{Peirce} This includes analogical, inductive, and abductive reasoning. Traditional knowledge bases tend to be brittle because a query posed in a slightly different way than the knowledge was entered will have no results. The vector space embedding allows some generalization of facts that helps overcome this brittleness. If an exact decomposition cannot be found, an approximate decomposition will be returned.

\subsection{Analogical Reasoning}
Semantic vectors are able to perform analogical reasoning with some simple arithmetic. This is also possible with vectors representing sets. For example, suppose we wanted to capture the meaning of the concept $Detroit$ in the sentence ``Detroit won the game on Monday night." The average of the word vectors $\{\vec{Detroit}$, $\vec{Detroit\_Lions}$, $\vec{Detroit\_Pistons}$, $\vec{Detroit\_Tigers}$, $\vec{Michigan}\}$, when $\vec{Cleveland}$ is added to it and $\vec{Detroit}$ subtracted from it, has the following as its highest weighted components: $\{\vec{Cleveland}$, $\vec{Cleveland\_Cavaliers}$, $\vec{Ohio}$, $\vec{Cavaliers}$, $\vec{Browns}$, $\vec{Cleveland\_Indians}\}$. This is a simple example, but illustrates how the notion of analogical reasoning can begin to be extended from individual words (where distributional semantic vectors have shown human levels of performance on multiple choice tests) to analogies between concepts. If we wish to know a more exact correspondence to each part of the analogy, we can use a word-level analogy over the much smaller dictionary created by the set of results of the set-level analogy.

\section{Learning from Definitions}
New terms can be added to the dictionary by defining them by means of the terms already in the dictionary along with the operations of union, intersection, negation, and analogy. For example, here is a definition of $nashi (tree)$\cite{randomhouse}:

\begin{itemize}
\item an Asian tree also cultivated in Australia and New Zealand, Pyrus pyrifolia, of the rose family, having apple-shaped, pear-colored, juicy fruit.
\end{itemize}

We can define a vector $\{trees\}$ representing the set of trees by adding together all known types of tree. If we have knowledge that some of these are $\{asian\_trees\}$, we can define a subset including just those. Otherwise, we need to define a set of $\{asian\_things\}$ and find the fuzzy intersection of that with $\{trees\}$. We can further intersect this with the set $\{rose\_family\}$. 

We now have a $\vec{nashi\_tree}$ vector that is a weighted sum of asian things, trees, and plants in the rose family. We can similarly create a $nashi\_fruit$ vector which is a weighted sum of apple-shape things, pear-colored things, juicy things, and fruit. But to represent that the fruit belongs to the plant is probably not best captured by intersection or union. Instead, we can define a relation vector $has\_fruit$ between a tree and its fruit by averaging the differences $(pear\;-\;pear\_tree)$, $(apple\;-\;apple\_tree)$ and so forth. Since we know that the $nashi\_tree$ and $nashi\_fruit$ vector have this same relationship, we can take $((nashi\_tree+has\_fruit) +(nashi\_fruit-has\_fruit))/2$ to get a revised $\vec{nashi\_tree}$ vector. $\vec{nashi\_tree}$ can now be added to the dictionary, and the phrase will be appropriately placed in the vector space to allow it to be used for analogical reasoning and similar operations.

\section{Conclusion}
Decomposing summed vectors is a reliable tool that allows high-dimensional vectors to be used to represent sets, probability distributions, ordered lists, and many other possibilities with exact solutions up to a certain size limit. The relationship between two concepts can also be decomposed into a set of known relations, forming a chain of deductive reasoning connecting the concepts. As opposed to working with symbolic knowledge bases, these vector-based methods degrade gracefully in the case of noise or missing information.

\end{document}